\documentclass[11pt,twoside,english]{article}
\usepackage[T1]{fontenc}
\usepackage[latin1]{inputenc}
\usepackage{color}
\usepackage{babel}
\usepackage{float}
\usepackage{bm}
\usepackage{amsmath}
\usepackage{amssymb}
\usepackage{graphicx}
\usepackage[authoryear]{natbib}
\usepackage[unicode=true,
 bookmarks=true,bookmarksnumbered=false,bookmarksopen=false,
 breaklinks=true,pdfborder={0 0 0},backref=false,colorlinks=true]
 {hyperref}
\usepackage{breakurl}

\makeatletter

\providecommand{\tabularnewline}{\\}
\floatstyle{ruled}
\newfloat{algorithm}{tbp}{loa}
\providecommand{\algorithmname}{Algorithm}
\floatname{algorithm}{\protect\algorithmname}

\usepackage{jmlr2e_modif}
\usepackage{algorithm}
\usepackage{algorithmic}

\makeatother

\begin{document}
\jmlrheading{}{}{}{}{}{Hannes Nickisch and Matthias Seeger}

\ShortHeadings{Multiple Kernel Learning: A Unifying Probabilistic
Viewpoint}{Nickisch and Seeger}

\title{Multiple Kernel Learning: A Unifying Probabilistic Viewpoint}

\author{\name Hannes Nickisch \email hannes@nickisch.org \\
 \addr Max Planck Institute for Intelligent Systems, Spemannstraße
38, 72076 Tübingen, Germany\\
\name Matthias Seeger \email matthias.seeger@epfl.ch \\
 \addr Ecole Polytechnique F\'{e}d\'{e}rale de Lausanne, INJ 339,
Station 14, 1015 Lausanne, Switzerland\\
}
\maketitle
\begin{abstract}
We present a probabilistic viewpoint to multiple kernel learning unifying
well-known regularised risk approaches and recent advances in approximate
Bayesian inference relaxations. The framework proposes a general objective
function suitable for regression, robust regression and classification
that is lower bound of the marginal likelihood and contains many regularised
risk approaches as special cases. Furthermore, we derive an efficient
and provably convergent optimisation algorithm.

\noindent \bigskip{}

\end{abstract}

\section{Introduction\label{sec:intro}}

Nonparametric kernel methods, cornerstones of machine learning today,
can be seen from different angles: as regularised risk minimisation
in function spaces \citep{Schoelkopf:02}, or as probabilistic Gaussian
process methods \citep{Rasmussen:06}. In these techniques, the kernel
(or equivalently covariance) function encodes interpolation characteristics
from observed to unseen points, and two basic statistical problems
have to be mastered. First, a latent function must be predicted which
fits data well, yet is as smooth as possible given the fixed kernel.
Second, the kernel function parameters have to be learned as well,
to best support predictions which are of primary interest. While the
first problem is simpler and has been addressed much more frequently
so far, the central role of learning the covariance function is well
acknowledged, and a substantial number of methods for ``learning
the kernel'', ``multiple kernel learning'', or ``evidence maximisation''
are available now. However, much of this work has firmly been associated
with one of the ``camps'' (referred to as \emph{regularised risk}
and \emph{probabilistic} in the sequel) with surprisingly little crosstalk
or acknowledgments of prior work across this boundary. In this paper,
we clarify the relationship between major regularised risk and probabilistic
kernel learning techniques precisely, pointing out advantages and
pitfalls of either, as well as algorithmic similarities leading to
novel powerful algorithms.

We develop a common analytical and algorithmical framework encompassing
approaches from both camps and provide clear insights into the optimisation
structure. Even though, most of the optimisation is non convex, we
show how to operate a provably convergent ``almost Newton'' method
nevertheless. Each step is not much more expensive than a gradient
based approach. Also, we do not require any foreign optimisation code
to be available. Our framework unifies kernel learning for regression,
robust regression and classification.

The paper is structured as follows: In section \ref{sec:frame}, we
introduce the regularised risk and the probabilistic view of kernel
learning. In increasing order of generality, we explain multiple kernel
learning (MKL, section \ref{sub:mkl}), maximum a posteriori estimation
(MAP, section \ref{sub:map}) and marginal likelihood maximisation
(MLM, section \ref{sub:mlm}). A taxonomy of the mutual relations
between the approaches and important special cases is given in section
\ref{sub:taxo}. Section \ref{sec:algos} introduces a general optimisation
scheme and section \ref{sec:conc} draws a conclusion.

\section{Kernel Methods and Kernel Learning\label{sec:frame}}

Kernel-based algorithms come in many shapes, however, the primary
goal is -- based on training data $\{(\mathbf{x}_{i},y_{i})\,|\, i=1..n\}$,
$\mathbf{x}_{i}\in{\cal X}$, $y_{i}\in\mathcal{Y}$ and a parametrised
kernel function $k_{\bm{\theta}}(\mathbf{x},\mathbf{x}')$ -- to predict
the output $y_{*}$ for unseen inputs $\mathbf{x}_{*}$. Often, linear
parametrisations $k_{\bm{\theta}}(\mathbf{x},\mathbf{x}')=\sum_{m=1}^{M}\theta_{m}k_{m}(\mathbf{x},\mathbf{x}')$
are used, where the $k_{m}$ are fixed positive definite functions,
and $\bm{\theta}\succeq\mathbf{0}$. Learning the kernel means finding
$\bm{\theta}$ to best support this goal. In general, kernel methods
employ a postulated latent function $u:{\cal X}\to\mathbb{R}$ whose
smoothness is controlled via the function space squared norm $\|u(\cdot)\|_{k_{\bm{\theta}}}^{2}$.
Most often, smoothness is traded against data fit, either enforced
by a \emph{loss function} $\ell(y_{i},u(\mathbf{x}_{i}))$ or modeled
by a \emph{likelihood} $\mathbb{P}(y_{i}|u_{i})$. Let us define kernel
matrices $\mathbf{K}_{\bm{\theta}}:=[k_{\bm{\theta}}(\mathbf{x}_{i},\mathbf{x}_{j})]_{ij}$,
and $\mathbf{K}_{m}:=[k_{m}(\mathbf{x}_{i},\mathbf{x}_{j})]_{ij}$
in $\mathbb{R}^{n\times n}$ and the vectors $\mathbf{y}:=[y_{i}]_{i}\in\mathcal{Y}^{n}$,
$\mathbf{u}:=[u(\mathbf{x}_{i})]_{i}\in\mathbb{R}^{n}$ collecting
outputs and latent function values, respectively.

The \emph{regularised risk} route to kernel prediction, which is followed
by any support vector machine (SVM) or ridge regression technique,
yields $\|u(\cdot)\|_{k_{\bm{\theta}}}^{2}+\frac{C}{n}\sum_{i=1}^{n}\ell(y_{i},u_{i})$
as criterion, enforcing smoothness of $u(\cdot)$ as well as good
data fit through th\emph{e }loss function $\frac{C}{n}\ell(y_{i},u(\mathbf{x}_{i}))$.
By the representer theorem, the minimiser can be written as $u(\cdot)=\sum_{i}\alpha_{i}k_{\bm{\theta}}(\cdot,\mathbf{x}_{i})$,
so that $\|u(\cdot)\|_{k_{\bm{\theta}}}^{2}=\bm{\alpha}^{\top}\mathbf{K}_{\bm{\theta}}\bm{\alpha}$
\citep{Schoelkopf:02}. As $\mathbf{u}=\mathbf{K}_{\bm{\theta}}\bm{\alpha}$,
the regularised risk problem is given by 
\begin{equation}
\min_{\mathbf{u}}\mathbf{u}^{\top}\mathbf{K}_{\bm{\theta}}^{-1}\mathbf{u}+\frac{C}{n}\sum_{i=1}^{n}\ell(y_{i},u_{i}).\label{eq:rkhs-estim}
\end{equation}

A \emph{probabilistic} viewpoint of the same setting is based on the
notion of a Gaussian process (GP) \citep{Rasmussen:06}: a Gaussian
random function $u(\cdot)$ with mean function $\mathbb{E}[u(\mathbf{x})]=m(\mathbf{x})\equiv0$
and covariance function $\mathbb{V}[u(\mathbf{x}),u(\mathbf{x}')]=\mathbb{E}[u(\mathbf{x})u(\mathbf{x}')]=k_{\bm{\theta}}(\mathbf{x},\mathbf{x}')$.
In practice, we only use finite-dimensional snapshots of the process
$u(\cdot)$: for example, $\mathbb{P}(\mathbf{u};\bm{\theta})=\mathcal{N}(\mathbf{u}|\mathbf{0},\mathbf{K}_{\bm{\theta}})$,
a zero-mean joint Gaussian with covariance matrix $\mathbf{K}_{\bm{\theta}}$.
We adopt this GP as prior distribution over $u(\cdot)$, estimating
the latent function as maximum of the posterior process $\mathbb{P}(u(\cdot)|\mathbf{y};\bm{\theta})\propto\mathbb{P}(\mathbf{y}|\mathbf{u})\mathbb{P}(u(\cdot);\bm{\theta})$.
Since the likelihood depends on $u(\cdot)$ only through the finite
subset $\{u(\mathbf{x}_{i})\}$, the posterior process has a finite-dimensional
representation: $\mathbb{P}(u(\cdot)|\mathbf{y},\mathbf{u})=\mathbb{P}(u(\cdot)|\mathbf{u})$,
so that $\mathbb{P}(u(\cdot)|\mathbf{y};\bm{\theta})=\int\mathbb{P}(u(\cdot)|\mathbf{u})\mathbb{P}(\mathbf{u}|\mathbf{y};\bm{\theta})\text{d}\mathbf{u}$
is specified by the joint distribution $\mathbb{P}(\mathbf{u}|\mathbf{y};\bm{\theta})$,
a probabilistic equivalent of the representer theorem. Kernel prediction
amounts to \emph{maximum a posteriori} (MAP) estimation 
\begin{equation}
\max\nolimits _{\mathbf{u}}\mathbb{P}(\mathbf{u}|\mathbf{y};\bm{\theta})\equiv\max\nolimits _{\mathbf{u}}\mathbb{P}(\mathbf{u};\bm{\theta})\mathbb{P}(\mathbf{y}|\mathbf{u})\equiv\min\nolimits _{\mathbf{u}}\mathbf{u}^{\top}\mathbf{K}_{\bm{\theta}}^{-1}\mathbf{u}-2\ln\mathbb{P}(\mathbf{y}|\mathbf{u})+\ln|\mathbf{K}_{\bm{\theta}}|,\label{eq:map-estim}
\end{equation}
ignoring an additive constant. Minimising equations \eqref{eq:rkhs-estim}
and \eqref{eq:map-estim} for any fixed kernel matrix $\mathbf{K}$
gives the same minimiser $\hat{\mathbf{u}}$ and prediction $u(\mathbf{x}_{*})=\hat{\mathbf{u}}^{\top}\mathbf{K}_{\bm{\theta}}^{-1}[k_{\bm{\theta}}(\mathbf{x}_{i},\mathbf{x}_{*})]_{i}$. 

The correspondence between likelihood and loss function bridges probabilistic
and regularised risk techniques. More specifically, any likelihood
$\mathbb{P}(\mathbf{y}|\mathbf{u})$ induces a loss function $\ell(\mathbf{y},\mathbf{u})$
via 
\[
-2\ln\mathbb{P}(\mathbf{y}|\mathbf{u})=-2\sum_{i}\ln\mathbb{P}(y_{i}|u_{i})\rightsquigarrow\frac{C}{n}\sum_{i=1}^{n}\ell(y_{i},u_{i})=\ell(\mathbf{y},\mathbf{u}),
\]
however some loss functions cannot be interpreted as a negative log
likelihood as shown in table \eqref{tab:loss-lik} and as discussed
for the SVM by \citet{Sollich:99a}. If, the likelihood is a \emph{log-concave}
function of $\mathbf{u}$, it corresponds to a convex loss function
\citep[Sect.~3.5.1]{Boyd:02}. Common loss functions and likelihoods
for classification $\mathcal{Y}=\{\pm1\}$ and regression $\mathcal{Y}=\mathbb{R}$
are listed in table \eqref{tab:loss-lik}.

\begin{table}
\begin{centering}
\resizebox{\textwidth}{!}{
\par\end{centering}

\begin{centering}
\begin{tabular}{|l||l|c||c|l|}
\hline 
$\mathcal{Y}$ & Loss function & $\ell(y_{i},u_{i})$ & $\mathbb{P}(y_{i}|u_{i})$ & Likelihood\tabularnewline
\hline 
\hline 
$\{\pm1\}$ & SVM Hinge loss & $\max(0,1-y_{i}u_{i})$ & \multicolumn{2}{c|}{$\nexists$}\tabularnewline
\hline 
$\{\pm1\}$ & Log loss & $\ln(\exp(-y_{i}u_{i})+1)$ & $1/(\exp(-\tau y_{i}u_{i})+1)$ & Logistic\tabularnewline
\hline 
\hline 
$\mathbb{R}$ & SVM $\epsilon$-insensitive loss & $\max(0,|y_{i}-u_{i}|/\epsilon-1)$ & \multicolumn{2}{c|}{$\nexists$}\tabularnewline
\hline 
$\mathbb{R}$ & Quadratic loss & $(y_{i}-u_{i})^{2}$ & $\mathcal{N}(y_{i}|u_{i},\sigma^{2})$ & Gaussian\tabularnewline
\hline 
$\mathbb{R}$ & Linear loss & $|y_{i}-u_{i}|$ & $\mathcal{L}(y_{i}|u_{i},\tau)$ & Laplace\tabularnewline
\hline 
\end{tabular}
\par\end{centering}

\begin{centering}
}
\par\end{centering}

\caption{\label{tab:loss-lik}Relations between loss functions and likelihoods}
\end{table}

In the following, we discuss several approaches to learn the kernel
parameters $\bm{\theta}$ and show how all of them can be understood
as instances of or approximations to Bayesian evidence maximisation.
Although the exposition MKL section \ref{sub:mkl} and MAP section
\ref{sub:map} use a linear parametrisation $\bm{\theta}\mapsto\mathbf{K}_{\bm{\theta}}=\sum_{m=1}^{M}\theta_{m}\mathbf{K}_{m}$,
much of the results in MLM \ref{sub:mlm} and all the aforementioned
discussion are still applicable to non-linear parametrisations.

\subsection{Multiple Kernel Learning\label{sub:mkl}}

A widely adopted regularised risk principle, known as \emph{multiple
kernel learning} (MKL) \citep{cristianini01kerneltarget,lanckriet04kernelSDP,bach04MKL},
is to minimise equation \eqref{eq:rkhs-estim} w.r.t. the kernel parameters
$\bm{\theta}$ as well. One obvious caveat is that for any fixed $\mathbf{u}$,
equation \eqref{eq:rkhs-estim} becomes ever smaller as $\theta_{m}\to\infty$:
it cannot per se play a meaningful statistical role. In order to prevent
this, researchers constrain the domain of $\bm{\theta}\in\bm{\Theta}$
and obtain 
\[
\min_{\bm{\theta}\in\bm{\Theta}}\min_{\mathbf{u}}\mathbf{u}^{\top}\mathbf{K}_{\bm{\theta}}^{-1}\mathbf{u}+\ell(\mathbf{y},\mathbf{u}),
\]
where $\bm{\Theta}=\{\bm{\theta}\succeq\mathbf{0},\:\|\bm{\theta}\|_{2}\le1\}$
or $\bm{\Theta}=\{\bm{\theta}\succeq\mathbf{0},\:\mathbf{1}^{\top}\bm{\theta}\le1\}$
\citep{varma07l1MKL}. Notably, these constraints are imposed independently
of the statistical problem, the model and of the parametrization $\bm{\theta}\mapsto\mathbf{K}_{\bm{\theta}}$.
The Lagrangian form of the MKL problem with parameter $\lambda$ and
a general $p$-norm unit ball constraint where $p\ge1$ \citep{kloft09lpMKL}
is given by 
\begin{equation}
\min_{\bm{\theta}\succeq\mathbf{0}}\phi_{\text{MKL}}(\bm{\theta}),\;\text{ where }\phi_{\text{MKL}}(\bm{\theta}):=\min_{\mathbf{u}}\mathbf{u}^{\top}\mathbf{K}_{\bm{\theta}}^{-1}\mathbf{u}+\ell(\mathbf{y},\mathbf{u})+\underbrace{\lambda\cdot\mathbf{1}^{\top}\bm{\theta}^{p}}_{\rho(\bm{\theta})},\;\lambda>0.\label{eq:mkl}
\end{equation}

Since, the \emph{regulariser} $\rho(\bm{\theta})$ for the kernel
parameter $\bm{\theta}$ is convex, the map $(\mathbf{u},\mathbf{K})\mapsto\mathbf{u}^{\top}\mathbf{K}^{-1}\mathbf{u}$
is jointly convex for $\mathbf{K}\succeq\mathbf{0}$ \citep{Boyd:02}
and the parametrisation $\bm{\theta}\mapsto\mathbf{K}_{\bm{\theta}}$
is linear, MKL is a jointly convex problem for $\bm{\theta}\succeq\mathbf{0}$
whenever the loss function $\ell(\mathbf{y},\mathbf{u})$ is convex.
Furthermore, there are efficient algorithms to solve equation \eqref{eq:mkl}
for large models \citep{sonnenburg06mkl}.

\subsection{Joint MAP Estimation\label{sub:map}}

\begin{figure}
\begin{centering}
\includegraphics[width=0.5\textwidth]{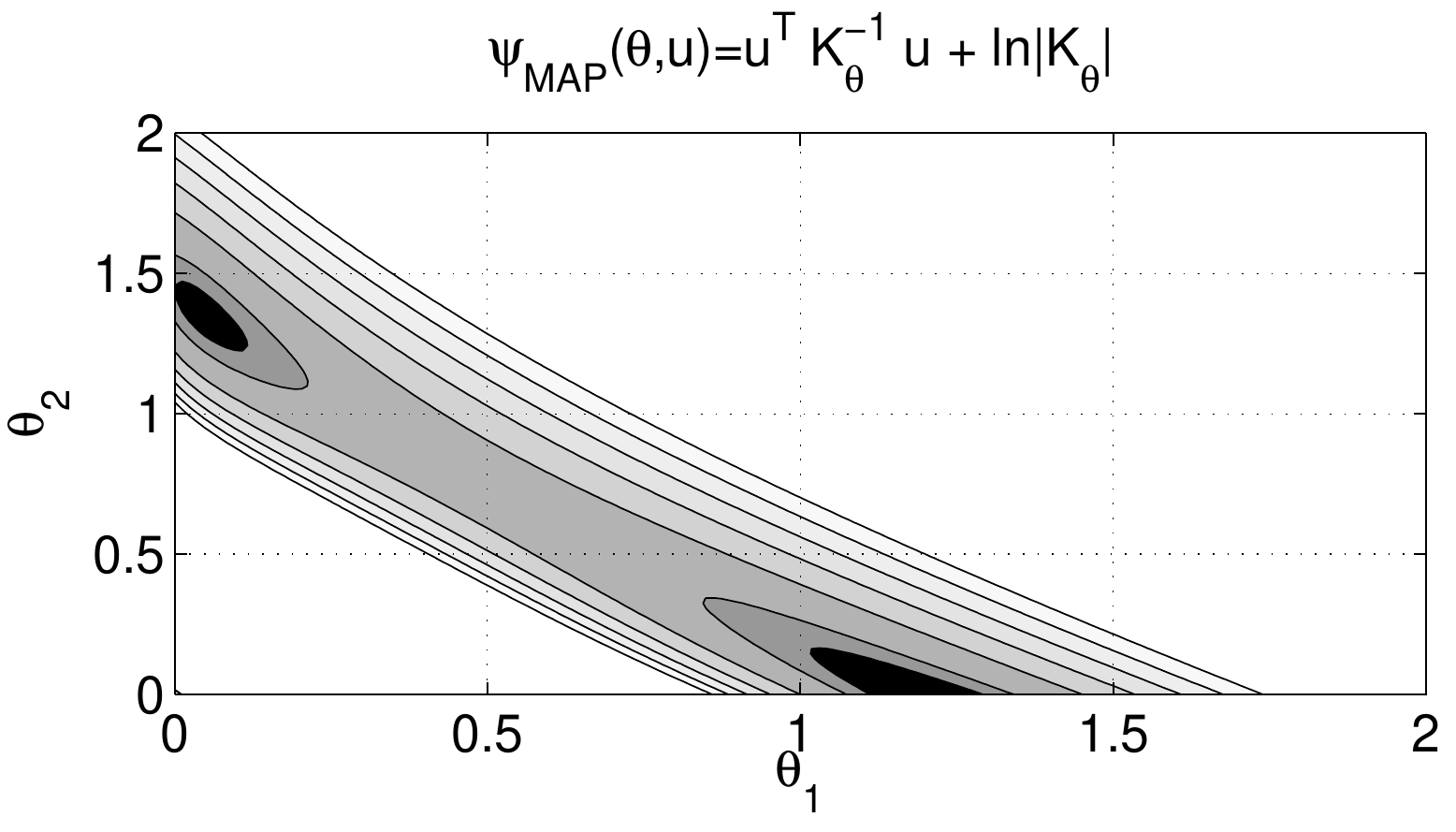}\hfill{}\includegraphics[width=0.5\textwidth]{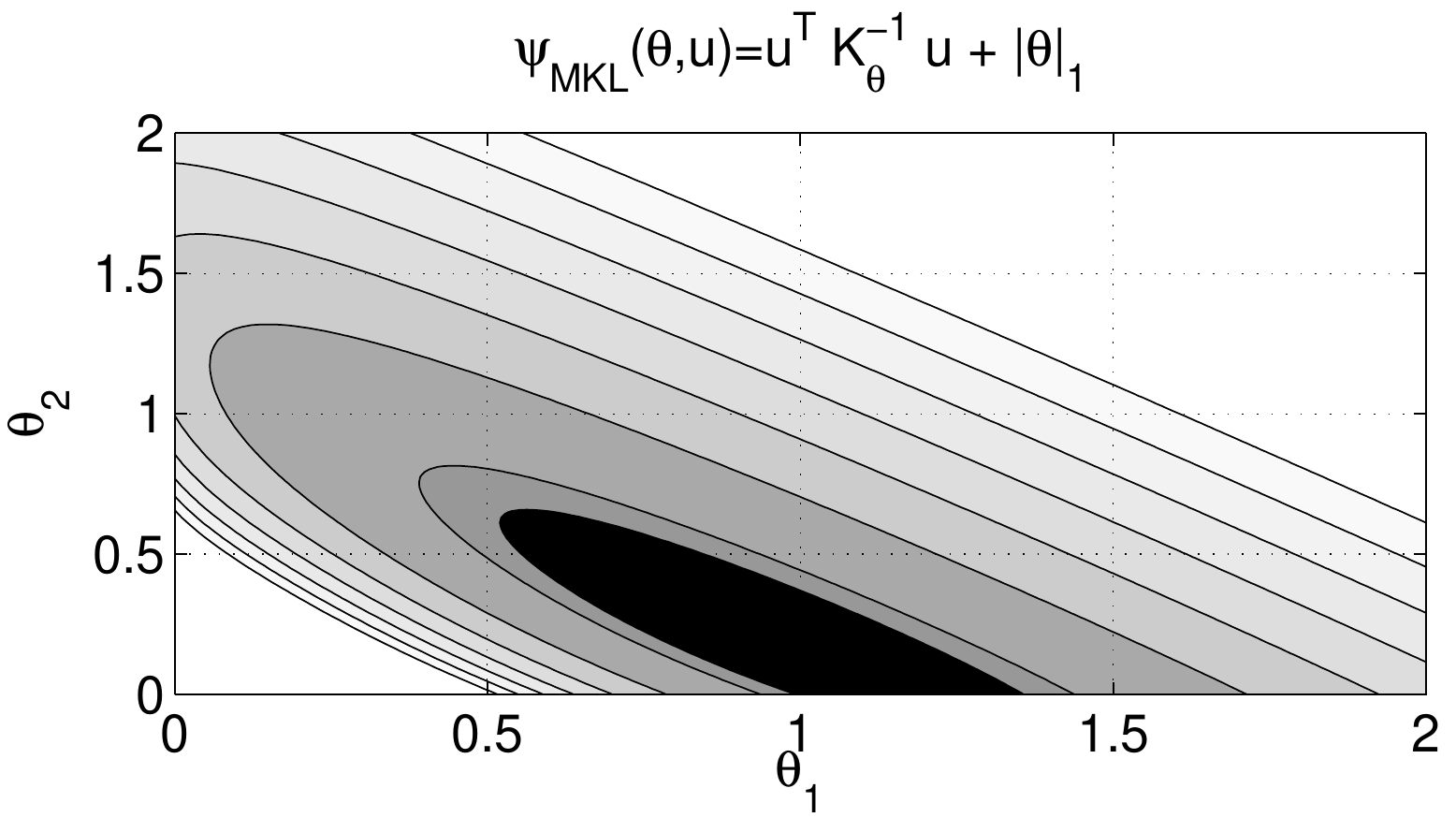}
\par\end{centering}

\caption{\label{fig:mkl-vs-map}Convex and non-convex building blocks of the
MKL and MAP objective function}
\end{figure}

Adopting a probabilistic MAP viewpoint, we can minimise equation \eqref{eq:map-estim}
w.r.t. $\mathbf{u}$ and $\bm{\theta}\succeq\mathbf{0}$: 
\begin{equation}
\min_{\bm{\theta}\succeq\mathbf{0}}\phi_{\text{MAP}}(\bm{\theta}),\;\text{ where }\phi_{\text{MAP}}(\bm{\theta}):=\min_{\mathbf{u}}\mathbf{u}^{\top}\mathbf{K}_{\bm{\theta}}^{-1}\mathbf{u}-2\ln\mathbb{P}(\mathbf{y}|\mathbf{u})+\ln|\mathbf{K}_{\bm{\theta}}|.\label{eq:map}
\end{equation}
While equation \eqref{eq:mkl} and equation \eqref{eq:map} share
the ``inner solution'' $\hat{\mathbf{u}}$ for fixed $\mathbf{K}_{\bm{\theta}}$
-- in case the loss $\ell(\mathbf{y},\mathbf{u})$ corresponds to
a likelihood $\mathbb{P}(\mathbf{y}|\mathbf{u})$ -- they are different
when it comes to optimising $\bm{\theta}$. The \emph{joint MAP} problem
is not in general jointly convex in $(\bm{\theta},\mathbf{u})$, since
$\bm{\theta}\mapsto\ln|\mathbf{K}_{\bm{\theta}}|$ is concave, see
figure \ref{fig:mkl-vs-map}. However, it is always a well-posed statistical
procedure, since $\ln|\mathbf{K}_{\bm{\theta}}|\to\infty$ as $\theta_{m}\to\infty$
for all $m$.

\begin{figure}
\begin{centering}
\includegraphics[width=1\textwidth]{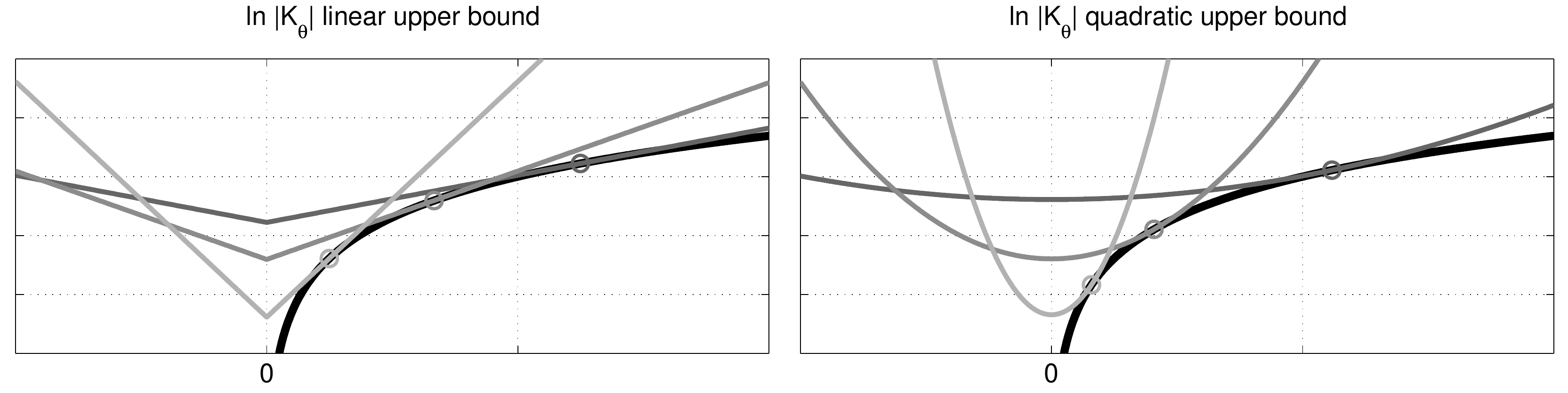}
\par\end{centering}

\caption{\label{fig:ld_ccv_bound}Convex upper bounds on (the concave non-decreasing)
$\ln|\mathbf{K}_{\bm{\theta}}|$}
By Fenchel duality, we can represent any concave non-decreasing function
and hence the log determinant function by $\ln|\mathbf{K}_{\bm{\theta}}|=\min_{\bm{\lambda}\succeq\mathbf{0}}\bm{\lambda}^{\top}|\bm{\theta}|^{p}-g^{*}(\bm{\lambda})$.
As a consequence, we obtain a piecewise polynomial upper bound for
any particular value $\bm{\lambda}$.
\end{figure}

We show in the following, how the regularisers $\rho(\bm{\theta})=\lambda\left\Vert \bm{\theta}\right\Vert _{p}^{p}$
of equation \eqref{eq:mkl} can be related to the probabilistic term
$f(\bm{\theta})=\ln|\mathbf{K}_{\bm{\theta}}|$. In fact, the same
reasoning can be applied to any concave non-decreasing function.

Since the function $\bm{\theta}\mapsto f(\bm{\theta})=\ln|\mathbf{K}_{\bm{\theta}}|$,
$\bm{\theta}\succeq\mathbf{0}$ is jointly concave, we can represent
it by $f(\bm{\theta})=\min_{\bm{\lambda}\succeq\mathbf{0}}\bm{\lambda}^{\top}\bm{\theta}-f^{*}(\bm{\lambda})$
where $f^{*}(\bm{\lambda})$ denotes Fenchel dual of $f(\bm{\theta})$.
Furthermore, the mapping $\bm{\vartheta}\mapsto\ln|\sum_{m=1}^{M}\sqrt[p]{\vartheta_{m}}\mathbf{K}_{m}|=f(\sqrt[p]{\bm{\vartheta}})=g(\bm{\vartheta})$,
$\bm{\vartheta}\succeq\mathbf{0}$ is jointly concave due to the composition
rule \citep[§3.2.4]{Boyd:02}, because $\bm{\vartheta}\mapsto\sqrt[p]{\bm{\vartheta}}$
is jointly concave and $\bm{\theta}\mapsto f(\bm{\theta})$ is non-decreasing
in all components $\theta_{m}$ as all matrices $\mathbf{K}_{m}$
are positive (semi-)definite which guarantees that the eigenvalues
of $\mathbf{K}_{\bm{\theta}}$ increase as $\theta_{m}$ increases.
Thus we can -- similarly to \citet{zhang10multistage} -- represent
$\ln|\mathbf{K}_{\bm{\theta}}|$ as
\[
\ln|\mathbf{K}_{\bm{\theta}}|=f(\bm{\theta})=g(\bm{\vartheta})=\min_{\bm{\lambda}\succeq\mathbf{0}}\bm{\lambda}^{\top}\bm{\vartheta}-g^{*}(\bm{\lambda})=\min_{\bm{\lambda}\succeq\mathbf{0}}\bm{\lambda}^{\top}|\bm{\theta}|^{p}-g^{*}(\bm{\lambda}).
\]
Choosing a particular value $\bm{\lambda}=\lambda\cdot\mathbf{1}$,
we obtain the bound $\ln|\mathbf{K}_{\bm{\theta}}|\le\lambda\cdot\left\Vert \bm{\theta}\right\Vert _{p}^{p}-g^{*}(\lambda\cdot\mathbf{1})$.
Figure \ref{fig:ld_ccv_bound} illustrates the bounds for $p=1$ and
$p=2$. The bottom line is that one can interpret the regularisers
$\rho(\bm{\theta})=\lambda\left\Vert \bm{\theta}\right\Vert _{p}^{p}$
in equation \eqref{eq:mkl} as corresponding to parametrised upper
bounds to the $\ln|\mathbf{K}_{\bm{\theta}}|$ part in equation \eqref{eq:map},
hence $\phi_{\text{MKL}}(\bm{\theta})=\psi_{\text{MAP}}(\bm{\theta},\bm{\lambda}=\lambda\cdot\mathbf{1})$,
where $\phi_{\text{MAP}}(\bm{\theta})=\min_{\bm{\lambda}\succeq\mathbf{0}}\psi_{\text{MAP}}(\bm{\theta},\bm{\lambda})$.
Far from an ad hoc choice to keep $\bm{\theta}$ small, the $\ln|\mathbf{K}_{\bm{\theta}}|$
term embodies the Occam's razor concept behind MAP estimation: overly
large $\bm{\theta}$ are ruled out, since their explanation of the
data $\mathbf{y}$ is extremely unlikely under the prior $\mathbb{P}(\mathbf{u};\bm{\theta})$.
The Occam's razor effect depends crucially on the proper normalization
of the prior \citep{MacKay:92}. For example, the weighting parameter
$C$ of $k$ ($k=C\tilde{k}$) can be learned by joint MAP: if $C=e^{c}$,
then equation \eqref{eq:map} is convex in $c$ for any fixed $\mathbf{u}$.
If kernel-regularised estimation equation \eqref{eq:rkhs-estim} is
interpreted as MAP estimation under a GP prior equation \eqref{eq:map-estim},
the correct extension to kernel learning is joint MAP: the MKL criterion
equation \eqref{eq:mkl} lacks prior normalization, which renders
MAP w.r.t. $\bm{\theta}$ meaningful in the first place. From a non-probabilistic
viewpoint, the $\ln|\mathbf{K}_{\bm{\theta}}|$ term comes with a
model and data dependent structure at least as complex as the rest
of equation \eqref{eq:mkl}.

While the MKL objective, equation \eqref{eq:mkl}, enjoys the benefit
of being convex in the (linear) kernel parameters $\bm{\theta}$,
this does not hold true for joint MAP estimation, equation \eqref{eq:map},
in general. We illustrate the differences in figure \ref{fig:mkl-vs-map}.
The function $\psi_{\text{MAP}}(\bm{\theta},\mathbf{u})$ is a building
block of the MAP objective $\phi_{\text{MAP}}(\bm{\theta})=\min_{\mathbf{u}}[\psi_{\text{MAP}}(\bm{\theta},\mathbf{u})-2\ln\mathbb{P}(\mathbf{y}|\mathbf{u})]$,
where
\[
\psi_{\text{MAP}}(\bm{\theta},\mathbf{u})=\underbrace{\mathbf{u}^{\top}\mathbf{K}_{\bm{\theta}}^{-1}\mathbf{u}}_{\psi_{\cup}(\bm{\theta},\mathbf{u})}+\underbrace{\ln|\mathbf{K}_{\bm{\theta}}|}_{\psi_{\cap}(\bm{\theta})}\le\psi_{\text{MKL}}(\bm{\theta},\mathbf{u})-g^{*}(\lambda\cdot\mathbf{1}),\:\psi_{\text{MKL}}(\bm{\theta},\mathbf{u})=\mathbf{u}^{\top}\mathbf{K}_{\bm{\theta}}^{-1}\mathbf{u}+\lambda\left\Vert \bm{\theta}\right\Vert _{p}^{p}.
\]
More concretely, $\psi_{\text{MAP}}(\bm{\theta},\mathbf{u})$ is a
sum of a nonnegative, jointly convex function $\psi_{\cup}(\bm{\theta},\mathbf{u})$
that is strictly decreasing in every component $\theta_{m}$ and a
concave function $\psi_{\cap}(\bm{\theta})$ that is strictly increasing
in every component $\theta_{m}$. Both functions $\psi_{\cup}(\bm{\theta},\mathbf{u})$
and $\psi_{\cap}(\bm{\theta})$ alone do not have a stationary point
due to their monotonicity in $\theta_{m}$. However, their sum can
have (even multiple) stationary points as shown in figure \ref{fig:mkl-vs-map}
on the left left. We can show, that the map $\mathbf{K}\mapsto\mathbf{u}^{\top}\mathbf{K}^{-1}\mathbf{u}+\ln|\mathbf{K}|$
is \emph{invex} i.e. every stationary point $\hat{\mathbf{K}}$ is
a global minimum. Using the convexity of $\mathbf{A}\mapsto\mathbf{u}^{\top}\mathbf{A}\mathbf{u}-\ln|\mathbf{A}|$
\citep{Boyd:02} and the fact that the derivative of $\mathbf{A}\mapsto\mathbf{A}^{-1}$
for $\mathbf{A}\in\mathbb{R}^{n\times n}$, $\mathbf{A}\succ\mathbf{0}$
has full rank $n^{2}$, we see by \citet[theorem 2.1]{mishra08invexOptim}
that $\mathbf{K}\mapsto\mathbf{u}^{\top}\mathbf{K}^{-1}\mathbf{u}+\ln|\mathbf{K}|$
is indeed invex.

Often, the MKL objective for the case $p=1$ is motivated by the fact
that the optimal solution $\bm{\theta}^{\star}$ is \emph{sparse}
\citep[e.g.][]{sonnenburg06mkl}, meaning that many components $\theta_{m}$
are zero. Figure \ref{fig:mkl-vs-map} illustrates that $\phi_{\text{MAP}}(\bm{\theta})$
also yields sparse solutions; in fact it enforces even more sparsity.
In MKL, $\phi_{\text{MAP}}(\bm{\theta})$ is simply relaxed to a convex
objective $\phi_{\text{MKL}}(\bm{\theta})$ at the expense of having
only a single less sparse solution. 

\subsubsection{Intuition for the Gaussian Case}

We can gain further intuition about the criteria $\phi_{\text{MKL}}$
and $\phi_{\text{MAP}}$ $ $by asking which \emph{matrices} $\mathbf{K}$
minimise them. For simplicity, assume that $\mathbb{P}(\mathbf{y}|\mathbf{u})=\mathcal{N}(\mathbf{y}|\mathbf{u},\sigma^{2}\mathbf{I})$
and $n/C=\sigma^{2}$, hence $\ell(\mathbf{y},\mathbf{u})=\frac{1}{\sigma^{2}}\left\Vert \mathbf{y}-\mathbf{u}\right\Vert _{2}^{2}$.
The inner minimiser $\hat{\mathbf{u}}$ for both $\phi_{\text{MKL}}$
and $\phi_{\text{MAP}}$ is given by $\mathbf{K}_{\bm{\theta}}^{-1}\hat{\mathbf{u}}=(\mathbf{K}_{\bm{\theta}}+\sigma^{2}\mathbf{I})^{-1}\mathbf{y}$.
With $\sigma^{2}\to0$, we find for joint MAP that $\frac{\partial}{\partial\mathbf{K}}\phi_{\text{MAP}}=\mathbf{0}$
results in $\hat{\mathbf{K}}=\mathbf{y}\mathbf{y}^{\top}$. While
this ``nonparametric'' estimate requires smoothing to be useful
in practice, closeness to $\mathbf{y}\mathbf{y}^{\top}$ is fundamental
to covariance estimation and can be found in regularised risk kernel
learning work \citep{cristianini01kerneltarget}. On the other hand,
for $\text{tr}(\mathbf{K}_{m})=1$ and hence $\rho(\bm{\theta})=\lambda\text{tr}(\mathbf{K}_{\bm{\theta}})=\lambda\left\Vert \bm{\theta}\right\Vert _{1}$,
$\frac{\partial}{\partial\mathbf{K}}\phi_{\text{MKL}}=\mathbf{0}$
leads to $\hat{\mathbf{K}}^{2}=\lambda^{-1}\mathbf{y}\mathbf{y}^{\top}$:
an odd way of estimating covariance, not supported by any statistical
literature we are aware of.

\subsection{Marginal Likelihood Maximisation\label{sub:mlm}}

While the joint MAP criterion uses a properly normalised prior distribution,
it is still not probabilistically consistent. Kernel learning amounts
to finding a value $\hat{\bm{\theta}}$ of high data likelihood, no
matter what the latent function $u(\cdot)$ is. The correct likelihood
to be maximised is \emph{marginal}: $\mathbb{P}(\mathbf{y}|\bm{\theta})=\int\mathbb{P}(\mathbf{y}|\mathbf{u})\mathbb{P}(\mathbf{u}|\bm{\theta})\text{d}\mathbf{u}$
(``max-sum''), while joint MAP employs the plug-in surrogate $\max_{\mathbf{u}}\mathbb{P}(\mathbf{y}|\mathbf{u})\mathbb{P}(\mathbf{u}|\bm{\theta})$
(``max-max''). \emph{Marginal likelihood maximisation} (MLM) is
also known as Bayesian estimation, and it underlies the EM algorithm
or maximum likelihood learning of conditional random fields just as
well: complexity is controlled (and overfitting avoided) by averaging
over unobserved variables $\mathbf{u}$ \citep{MacKay:92}, rather
than plugging in some point estimate $\hat{\mathbf{u}}$

\begin{equation}
\phi_{\text{MLM}}(\bm{\theta}):=-2\ln\int\mathcal{N}(\mathbf{u}|\mathbf{0},\mathbf{K}_{\bm{\theta}})\mathbb{P}(\mathbf{y}|\mathbf{u})\text{d}\mathbf{u}.\label{eq:mlm}
\end{equation}

\subsubsection{The Gaussian Case}

Before developing a general MLM approximation, we note an important
analytically solvable exception: for Gaussian likelihood $\mathbb{P}(\mathbf{y}|\mathbf{u})=\mathcal{N}(\mathbf{y}|\mathbf{u},\sigma^{2}\mathbf{I})$,
$\mathbb{P}(\mathbf{y}|\bm{\theta})=\mathcal{N}(\mathbf{y}|\mathbf{0},\mathbf{K}_{\bm{\theta}}+\sigma^{2}\mathbf{I})$,
and MLM becomes 
\begin{equation}
\phi_{\text{GAU}}(\bm{\theta}):=\mathbf{y}^{\top}(\mathbf{K}_{\bm{\theta}}+\sigma^{2}\mathbf{I})^{-1}\mathbf{y}+\ln|\mathbf{K}_{\bm{\theta}}+\sigma^{2}\mathbf{I}|.\label{eq:mlm-gauss}
\end{equation}
Even if the primary purpose is classification, the Gaussian likelihood
is used for its analytical simplicity \citep{kapoor09GPobjectCat}.
Only for the Gaussian case, joint MAP and MLM have an analytically
closed form. From the product formula of Gaussians \citep[§5.1]{brookes05MRM}

\[
\mathbb{Q}(\mathbf{u}):=\mathcal{N}(\mathbf{u}|\mathbf{0},\mathbf{K}_{\bm{\theta}})\mathcal{N}(\mathbf{y}|\mathbf{u},\bm{\Gamma})=\mathcal{N}(\mathbf{y}|\mathbf{0},\mathbf{K}_{\bm{\theta}}+\bm{\Gamma})\mathcal{N}(\mathbf{u}|\mathbf{m},\mathbf{V}),
\]
where $\mathbf{V}=(\mathbf{K}_{\bm{\theta}}^{-1}+\bm{\Gamma}^{-1})^{-1}$
and $\mathbf{m}=\mathbf{V}\bm{\Gamma}^{-1}\mathbf{y}$ we can deduce
that
\begin{equation}
-2\ln\int\mathbb{Q}(\mathbf{u})\text{d}\mathbf{u}=\ln|\mathbf{K}_{\bm{\theta}}^{-1}+\bm{\Gamma}^{-1}|+\min_{\mathbf{u}}[-2\ln\mathbb{Q}(\mathbf{u})]-n\ln|2\pi|.\label{eq:int-max}
\end{equation}
Using $\sigma^{2}\mathbf{I}=\bm{\Gamma}$ and $\min_{\mathbf{u}}[-2\ln\mathbb{Q}(\mathbf{u})]=-2\ln\mathbb{Q}(\mathbf{m})$,
we see that by
\begin{equation}
\phi_{\text{MAP/GAU}}(\bm{\theta}):\stackrel{\text{c}}{=}\phi_{\text{GAU}}(\bm{\theta})-\ln|\mathbf{K}_{\bm{\theta}}^{-1}+\sigma^{-2}\mathbf{I}|\stackrel{\text{c}}{=}\mathbf{y}^{\top}(\mathbf{K}_{\bm{\theta}}+\sigma^{2}\mathbf{I})^{-1}\mathbf{y}+\ln|\mathbf{K}_{\bm{\theta}}|\label{eq:map-gauss}
\end{equation}
MLM and MAP are very similar for the Gaussian case.

The ``ridge regression'' approximation is also used together with
$p$-norm constraints instead of the $\ln|\mathbf{K}_{\bm{\theta}}|$
term \citep{cortes09l2MKL} 
\begin{equation}
\phi_{\text{RR}}(\bm{\theta}):=\mathbf{y}^{\top}(\mathbf{K}_{\bm{\theta}}+\sigma^{2}\mathbf{I})^{-1}\mathbf{y}+\lambda\left\Vert \bm{\theta}\right\Vert _{p}^{p}.\label{eq:mkl-gauss}
\end{equation}
Unfortunately, most GP methods to date work with a Gaussian likelihood
for simplicity, a restriction which often proves short-sighted. Gaussian-linear
models come with unrealistic properties, and benefits of MLM over
joint MAP cannot be realised.

Kernel parameter learning has been an integral part of probabilistic
GP methods from the very beginning. \citet{williams96GPR} proposed
MLM for Gaussian noise equation \ref{eq:mlm-gauss}, fifteen years
ago. They treated sums of exponential and linear kernels as well as
learning lengthscales (ARD), predating recent proposals such as ``products
of kernels'' \citep{varma09generalMKL}.

\subsubsection{The General Case}

In general, joint MAP always has the analytical form equation \ref{eq:map},
while $\mathbb{P}(\mathbf{y}|\bm{\text{\ensuremath{\theta}}})$ can
only be approximated. For non-Gaussian $\mathbb{P}(\mathbf{y}|\mathbf{u})$,
numerous approximate inference methods have been proposed, specifically
motivated by learning kernel parameters via MLM. The simplest such
method is Laplace's approximation, applied to GP binary and multi-way
classification by \citet{Williams:98b}: starting with convex joint
MAP, $\ln\mathbb{P}(\mathbf{y},\mathbf{u})$ is expanded to second
order around the posterior mode $\hat{\mathbf{u}}$. More recent approximations
\citet{girolami05MKLhierBayes,girolami06GPDataIntegration} can be
much more accurate, yet come with non-convex problems and less robust
algorithms \citep{Nickisch:08}. In this paper, we concentrate on
the variational lower bound relaxation (VB) by \citet{Jaakkola:00},
which is convex for log-concave likelihoods $\mathbb{P}(\mathbf{y}|\mathbf{u})$
\citep{nickisch09convexInf}, providing a novel simple and efficient
algorithm. While our VB approximation to MLM is more expensive to
run than joint MAP for non-Gaussian likelihood (even using Laplace's
approximation), the implementation complexity of our VB algorithm
is comparable to what is required in the Gaussian noise case equation
\ref{eq:mlm-gauss}.

More, specifically, we exploit that super-Gaussian of likelihoods
$\mathbb{P}(y_{i}|u_{i})$ can be lower bounded by scaled Gaussians
$\mathcal{N}_{\gamma_{i}}$ of any width $\gamma_{i}$:
\[
\mathbb{P}(y_{i}|u_{i})=\max_{\gamma_{i}>0}\mathcal{N}_{\gamma_{i}}=\max_{\gamma_{i}>0}\exp\left(\beta_{i}u_{i}-\frac{u_{i}^{2}}{2\gamma_{i}}-\frac{1}{2}h_{i}(\gamma_{i})\right),
\]
where $\beta_{i}\propto y_{i}$ are constants, and $h_{i}(\cdot)$
is convex \citep{nickisch09convexInf} whenever the likelihood $\mathbb{P}(y_{i}|u_{i})$
is log-concave. If the posterior distribution is $\mathbb{P}(\mathbf{u}|\mathbf{y})=Z^{-1}\mathbb{P}(\mathbf{y}|\mathbf{u})\mathbb{P}(\mathbf{u})$,
then $\ln Z\ge Ce^{-\psi_{\text{VB}}(\bm{\theta},\bm{\gamma})/2}$
by plugging in these bounds, where $C$ is a constant and 
\begin{equation}
\phi_{\text{VB}}(\bm{\theta}):=\min_{\bm{\gamma}\succ\mathbf{0}}\psi_{\text{VB}}(\bm{\theta},\bm{\gamma}),\quad\psi_{\text{VB}}(\bm{\theta},\bm{\gamma}):=h(\bm{\gamma})-2\ln\int\mathcal{N}(\mathbf{u}|\mathbf{0},\mathbf{K}_{\bm{\theta}})e^{\mathbf{u}^{\top}(\bm{\beta}-\frac{1}{2}\bm{\Gamma}{}^{-1}\mathbf{u})}\text{d}\mathbf{u},\label{eq:vb-int}
\end{equation}
$h(\bm{\gamma}):=\sum_{i}h_{i}(\gamma_{i})$, $\bm{\Gamma}:=\text{dg}(\bm{\gamma})$.
The variational relaxation%
\footnote{Generalisations to other super-Gaussian potentials (log-concave or
not) or models including linear couplings and mixed potentials are
given by \citet{nickisch09convexInf}.%
} amounts to maximising the lower bound, which means that $\mathbb{P}(\mathbf{u}|\mathbf{y})$
is fitted by the \emph{Gaussian} approximation $\mathbb{Q}(\mathbf{u}|\mathbf{y};\bm{\gamma})$
with covariance matrix $\mathbf{V}=(\mathbf{K}_{\bm{\theta}}^{-1}+\bm{\Gamma}{}^{-1})^{-1}$
\citep{nickisch09convexInf}. Alternatively, we can interpret $\psi_{\text{VB}}(\bm{\theta},\bm{\gamma})$
as an upper bound on the Kullback-Leibler divergence $\text{KL}(\mathbb{Q}(\mathbf{u}|\mathbf{y};\bm{\gamma})||\mathbb{P}(\mathbf{u}|\mathbf{y}))$
\citep[§2.5.9]{nickisch10phd}, a measure for the dissimilarity between
the exact posterior $\mathbb{P}(\mathbf{u}|\mathbf{y})$ and the parametrised
Gaussian approximation $\mathbb{Q}(\mathbf{u}|\mathbf{y};\bm{\gamma})$.

Finally, note that by equation \eqref{eq:int-max}, $\psi_{\text{VB}}(\bm{\theta},\bm{\gamma})$
can also be written as
\begin{equation}
\psi_{\text{VB}}(\bm{\theta},\bm{\gamma})=\ln|\mathbf{K}_{\bm{\theta}}^{-1}+\bm{\Gamma}^{-1}|+h(\bm{\gamma})+\min_{\mathbf{u}}R(\mathbf{u},\bm{\theta},\bm{\gamma})+\ln|\mathbf{K}_{\bm{\theta}}|,\label{eq:vb-max-ga}
\end{equation}
where $R(\mathbf{u},\bm{\theta},\bm{\gamma})=\mathbf{u}^{\top}(\mathbf{K}_{\bm{\theta}}^{-1}+\bm{\Gamma}^{-1})\mathbf{u}-2\bm{\beta}^{\top}\mathbf{u}$.
Using the concavity of $\bm{\gamma}^{-1}\mapsto\ln|\mathbf{K}_{\bm{\theta}}^{-1}+\bm{\Gamma}^{-1}|$
and Fenchel duality $\ln|\mathbf{K}_{\bm{\theta}}^{-1}+\bm{\Gamma}^{-1}|=\min_{\mathbf{z}\succ\mathbf{0}}\mathbf{z}^{\top}\bm{\gamma}^{-1}-g_{\bm{\theta}}^{*}(\mathbf{z})=\hat{\mathbf{z}}_{\bm{\theta}}^{\top}\bm{\gamma}^{-1}-g_{\bm{\theta}}^{*}(\hat{\mathbf{z}}_{\bm{\theta}})$,
with the optimal value $\hat{\mathbf{z}}_{\bm{\theta}}=\text{dg}(\mathbf{V})$,
we can reformulate $\psi_{\text{VB}}(\bm{\theta},\bm{\gamma})$ as
\[
\psi_{\text{VB}}(\bm{\theta},\bm{\gamma})=\min_{\mathbf{z}\succ\mathbf{0}}[\mathbf{z}^{\top}\bm{\gamma}^{-1}-g_{\bm{\theta}}^{*}(\mathbf{z})]+h(\bm{\gamma})+\min_{\mathbf{u}}R(\mathbf{u},\bm{\theta},\bm{\gamma})+\ln|\mathbf{K}_{\bm{\theta}}|,
\]
which allows to perform the minimisation w.r.t. $\bm{\gamma}$ in
closed form \citep[§3.5.6]{nickisch10phd}:

\begin{equation}
\phi_{\text{VB}}(\bm{\theta})=\min_{\mathbf{z}\succ\mathbf{0}}\psi_{\text{VB}}(\bm{\theta},\mathbf{z}),\quad\psi_{\text{VB}}(\bm{\theta},\mathbf{z})=\min_{\mathbf{u}}\mathbf{u}^{\top}\mathbf{K}_{\bm{\theta}}^{-1}\mathbf{u}+\tilde{\ell}_{\mathbf{z}}(\mathbf{y},\mathbf{u})-g_{\bm{\theta}}^{*}(\mathbf{z})+\ln|\mathbf{K}_{\bm{\theta}}|,\label{eq:vb-max-z}
\end{equation}
where $\tilde{\ell}_{\mathbf{z}}(\mathbf{y},\mathbf{u}):=2\bm{\beta}^{\top}(\mathbf{v}-\mathbf{u})-2\ln\mathbb{P}(\mathbf{y}|\mathbf{v})$
and finally $\mathbf{v}=\text{sign}(\mathbf{u})\odot\sqrt{\mathbf{u}^{2}+\mathbf{z}}$.
Note that for $\mathbf{z}=\mathbf{0}$, we exactly recover joint MAP
estimation, equation \eqref{eq:map}, as $\mathbf{z}=\mathbf{0}$
implies $\mathbf{u}=\mathbf{v}$ and $\tilde{\ell}_{\mathbf{z}}(\mathbf{y},\mathbf{u})=\ell(\mathbf{y},\mathbf{u})$.
For fixed $\bm{\theta}$, the optimal value $\hat{\mathbf{z}}_{\bm{\theta}}=\text{dg}(\mathbf{V})$
corresponds to the marginal variances of the Gaussian approximation
$\mathbb{Q}(\mathbf{u}|\mathbf{y};\bm{\gamma})$: Variational inference
corresponds to variance-smoothed joint MAP estimation \citep{nickisch10phd}
with a loss function $\tilde{\ell}(\mathbf{y},\mathbf{u},\bm{\theta})$
that explicitly depends on the kernel parameters $\bm{\theta}$. We
have two equivalent representations of the loss $\tilde{\ell}(\mathbf{y},\mathbf{u},\bm{\theta})$
that directly follow from equations \eqref{eq:vb-max-ga} and \eqref{eq:vb-max-z}:
\begin{eqnarray*}
\tilde{\ell}(\mathbf{y},\mathbf{u},\bm{\theta}) & = & \min_{\bm{\gamma}\succ\mathbf{0}}[\ln|\mathbf{K}_{\bm{\theta}}^{-1}+\bm{\Gamma}^{-1}|+h(\bm{\gamma})+\mathbf{u}^{\top}\bm{\Gamma}^{-1}\mathbf{u}-2\bm{\beta}^{\top}\mathbf{u}],\:\text{and}\\
\tilde{\ell}(\mathbf{y},\mathbf{u},\bm{\theta}) & = & \min_{\mathbf{z}\succ\mathbf{0}}[2\bm{\beta}^{\top}(\mathbf{v}-\mathbf{u})-2\ln\mathbb{P}(\mathbf{y}|\mathbf{v})-g_{\bm{\theta}}^{*}(\mathbf{z})],\;\mathbf{v}=\text{sign}(\mathbf{u})\odot\sqrt{\mathbf{u}^{2}+\mathbf{z}}.
\end{eqnarray*}
Our VB problem is $\min_{\bm{\theta}\succeq\mathbf{0},\bm{\gamma}\succ\mathbf{0}}\psi_{\text{VB}}(\bm{\theta},\bm{\gamma})$
or equivalently $\min_{\bm{\theta}\succeq\mathbf{0},\mathbf{z}\succ\mathbf{0}}\psi_{\text{VB}}(\bm{\theta},\mathbf{z})$.
The inner variables here are $\bm{\gamma}$ and $\mathbf{z}$, in
addition to $\mathbf{u}$ in joint MAP. There are further similarities:
since $\psi_{\text{VB}}(\bm{\theta},\bm{\gamma})=-2\ln\int e^{-R(\mathbf{u},\bm{\gamma},\bm{\theta})}\text{d}\mathbf{u}+h(\bm{\gamma})+\ln|2\pi\mathbf{K}_{\bm{\theta}}|$,
$(\bm{\gamma},\bm{\theta})\mapsto\psi_{\text{VB}}-\ln|\mathbf{K}_{\bm{\theta}}|$
is jointly convex for $\bm{\gamma}\succ\mathbf{0}$, $\bm{\theta}\succeq\mathbf{0}$,
by the joint convexity of $(\mathbf{u},\bm{\gamma},\bm{\theta})\mapsto R$
and Prékopa's theorem \citep[§3.5.2]{Boyd:02}. Joint MAP and VB share
the same convexity structure. In contrast, approximating $\mathbb{P}(\mathbf{y}|\bm{\theta})$
by other techniques like Expectation Propagation \citep{Minka:01a}
or general Variational Bayes \citep{Opper:09} does not even constitute
convex problems for fixed $\bm{\theta}$.

\subsection{Summary and Taxonomy\label{sub:taxo}}

\begin{table}[h]
\begin{centering}
\resizebox{\textwidth}{!}{
\par\end{centering}

\begin{centering}
\begin{tabular}{|l|l|}
\hline 
Name & Objective function\tabularnewline
\hline 
\hline 
Marginal Likelihood Maximisation & $\phi_{\text{MLM}}(\bm{\theta})=-2\ln\left[\int\mathcal{N}(\mathbf{u}|\mathbf{0},\mathbf{K}_{\bm{\theta}})\mathbb{P}(\mathbf{y}|\mathbf{u})\text{d}\mathbf{u}\right]$\tabularnewline
\hline 
Variational Bounds  & $\phi_{\text{VB}}(\bm{\theta})=\min_{\bm{\gamma}\succ\mathbf{0}}\psi_{\text{VB}}(\bm{\theta},\bm{\gamma})\ge\phi_{\text{MLM}}(\bm{\theta})$
by $\mathbb{P}(y_{i}|u_{i})\ge\mathcal{N}_{\gamma_{i}}$\tabularnewline
\hline 
Maximum A Posteriori & $\phi_{\text{MAP}}(\bm{\theta})=-2\ln\left[\max_{\mathbf{u}}\mathcal{N}(\mathbf{u}|\mathbf{0},\mathbf{K}_{\bm{\theta}})\mathbb{P}(\mathbf{y}|\mathbf{u})\right]=\psi_{\text{VB}}(\bm{\theta},\mathbf{z}=\mathbf{0})$\tabularnewline
\hline 
Multiple Kernel Learning & $\phi_{\text{MKL}}(\bm{\theta})=\phi_{\text{MAP}}(\bm{\theta})+\lambda\left\Vert \bm{\theta}\right\Vert _{p}^{p}-\ln|\mathbf{K}_{\bm{\theta}}|=\psi_{\text{MAP}}(\bm{\theta},\bm{\lambda}=\lambda\cdot\mathbf{1})$\tabularnewline
\hline 
\end{tabular}
\par\end{centering}

\begin{centering}
}
\par\end{centering}

\begin{centering}
\bigskip{}

\par\end{centering}

\begin{centering}
\resizebox{\textwidth}{!}{
\par\end{centering}

\begin{centering}
\begin{tabular}{r|l|c|l|l}
\multicolumn{1}{r}{} & \multicolumn{1}{l}{\hspace{3mm}General $\mathbb{P}(y_{i}|u_{i})$} & \multicolumn{1}{c}{} & \multicolumn{1}{l}{\hspace{3mm}Gaussian $\mathbb{P}(y_{i}|u_{i})$} & \tabularnewline
\cline{2-2} \cline{4-4} 
 & $\phi_{\text{MLM}}(\bm{\theta})$, eq. \eqref{eq:mlm} & $\longrightarrow$ & $\phi_{\text{GAU}}(\bm{\theta})$, eq. \eqref{eq:mlm-gauss} & \tabularnewline[1mm]
\cline{2-2} 
\multicolumn{1}{r}{Super-Gaussian Bounding} & \multicolumn{1}{l}{\hspace{15mm}$\downarrow$} &  & \hspace{15mm}$\downarrow$ & Bound is tight\tabularnewline
\cline{2-2} 
 & $\phi_{\text{VB}}(\bm{\theta})$, eq. \eqref{eq:vb-int} & $\longrightarrow$ & $\phi_{\text{GAU}}(\bm{\theta})$, eq. \eqref{eq:mlm-gauss} & \tabularnewline[1mm]
\cline{2-2} \cline{4-4} 
\multicolumn{1}{r}{Maximum instead of integral} & \multicolumn{1}{l}{\hspace{15mm}$\downarrow$} & \multicolumn{1}{c}{} & \multicolumn{1}{l}{\hspace{15mm}$\downarrow$} & Mode $\equiv$ mean\tabularnewline
\cline{2-2} \cline{4-4} 
 & $\phi_{\text{MAP}}(\bm{\theta})$, eq. \eqref{eq:map} & $\longrightarrow$ & $\phi_{\text{MAP/GAU}}(\bm{\theta})$, eq. \eqref{eq:map-gauss} & \tabularnewline[1mm]
\cline{2-2} \cline{4-4} 
\multicolumn{1}{r}{Bound $\ln|\mathbf{K}_{\bm{\theta}}|\le\lambda\left\Vert \bm{\theta}\right\Vert _{p}^{p}-g^{*}(\lambda\mathbf{1})$ } & \multicolumn{1}{l}{\hspace{15mm}$\downarrow$} & \multicolumn{1}{c}{} & \multicolumn{1}{l}{\hspace{15mm}$\downarrow$} & \tabularnewline
\cline{2-2} \cline{4-4} 
 & $\phi_{\text{MKL}}(\bm{\theta})$, eq. \eqref{eq:mkl} & $\longrightarrow$ & $\phi_{\text{RR}}(\bm{\theta})$, eq. \eqref{eq:mkl-gauss} & \tabularnewline[1mm]
\cline{2-2} \cline{4-4} 
\end{tabular}
\par\end{centering}

\begin{centering}
}
\par\end{centering}

\caption{\label{tab:objectives}Taxonomy of kernel learning objective functions}

The upper table visualises the relationship between several kernel
learning objective functions for arbitrary likelihood/loss functions:
Marginal likelihood maximisation (MLM) can be bounded by variational
bounds (VB) and maximum a posteriori estimation (MAP) is a special
case $\mathbf{z}=\mathbf{0}$ thereof. Finally multiple kernel learning
(MKL) can be understood as an upper bound to the MAP estimation objective
$\bm{\lambda}=\lambda\cdot\mathbf{1}$. The lower table complements
the upper table by also covering the analytically important Gaussian
case.
\end{table}

In the last paragraphs, we have detailed how a variety of kernel learning
approaches can be obtained from Bayesian marginal likelihood maximisation
in a sequence of nested upper bounding steps. Table \ref{tab:objectives}
nicely illustrates how many kernel learning objectives are related
to each other -- either by upper bounds or by Gaussianity assumptions.
We can clearly see, that $\phi_{\text{VB}}(\bm{\theta})$ -- as an
upper bound to the negative log marginal likelihood -- can be seen
as the mother function. For a special case, $\mathbf{z}=\mathbf{0}$,
we obtain joint maximum a posteriori estimation, where the loss functions
does not depend on the kernel parameters. Going further, a particular
instance $\bm{\lambda}=\lambda\cdot\mathbf{1}$ yields the widely
use multiple kernel learning objective that becomes convex in the
kernel parameters $\bm{\theta}$. In the following, we will concentrate
on the optimisation and computational similarities between the approaches.

\section{Algorithms\label{sec:algos}}

In this section, we derive a simple, provably convergent and efficient
algorithm for MKL, joint MAP and VB. We use the Lagrangian form of
equation \eqref{eq:mkl} and $\ell(\mathbf{y},\mathbf{u}):=-2\ln\mathbb{P}(\mathbf{y}|\mathbf{u})$:
\begin{eqnarray*}
\psi_{\text{MKL}}(\bm{\theta},\mathbf{u}) & = & \mathbf{u}^{\top}\mathbf{K}^{-1}\mathbf{u}+\qquad\;\ell(\mathbf{y},\mathbf{u})\qquad\qquad\qquad\qquad\qquad\;\,+\lambda\cdot\mathbf{1}^{\top}\bm{\theta},\:\lambda>0,\\
\psi_{\text{MAP}}(\bm{\theta},\mathbf{u}) & = & \mathbf{u}^{\top}\mathbf{K}_{\bm{\theta}}^{-1}\mathbf{u}+\qquad\;\ell(\mathbf{y},\mathbf{u})\qquad\qquad\qquad\qquad\qquad\;\,+\ln|\mathbf{K}_{\bm{\theta}}|,\quad\text{and}\\
\psi_{\text{VB}}(\bm{\theta},\mathbf{u}) & = & \mathbf{u}^{\top}\mathbf{K}_{\bm{\theta}}^{-1}\mathbf{u}+\min_{\mathbf{z}\succ\mathbf{0}}\left[\ell(\mathbf{y},\mathbf{v})+2\bm{\beta}^{\top}(\mathbf{v}-\mathbf{u})-g_{\bm{\theta}}^{*}(\mathbf{z})\right]+\ln|\mathbf{K}_{\bm{\theta}}|,\\
 &  & \qquad\qquad\qquad\qquad\qquad\qquad\qquad\qquad\quad\text{where}\;\mathbf{v}=\text{sign}(\mathbf{u})\odot\sqrt{\mathbf{u}^{2}+\mathbf{z}}.
\end{eqnarray*}
Many previous algorithms use alternating minimization, which is easy
to implement but tends to converge slowly. Both $\phi_{\text{VB}}$
and $\phi_{\text{MAP}}$ are jointly convex up to the concave $\bm{\theta}\mapsto\ln|\mathbf{K}_{\bm{\theta}}|$
part. Since $\ln|\mathbf{K}_{\bm{\theta}}|=\min_{\bm{\lambda}\succ\mathbf{0}}\bm{\lambda}^{\top}\bm{\theta}-f^{*}(\bm{\lambda})$
\citep[Legendre duality,][]{Boyd:02}, joint MAP becomes $\min_{\bm{\lambda}\succ\mathbf{0},\bm{\theta}\succeq\mathbf{0},\mathbf{u}}\phi_{\bm{\lambda}}(\bm{\theta},\mathbf{u})$
with $\phi_{\bm{\lambda}}:=\mathbf{u}^{\top}\mathbf{K}_{\bm{\theta}}^{-1}\mathbf{u}+\ell(\mathbf{y},\mathbf{u})+\bm{\lambda}^{\top}\bm{\theta}-f^{*}(\bm{\lambda})$
which is jointly convex in $(\bm{\theta},\mathbf{u})$. Algorithm
\ref{alg:double-loop} iterates between refits of $\bm{\lambda}$
and joint Newton updates of $(\bm{\theta},\mathbf{u})$.

\begin{algorithm}[ht]
\begin{algorithmic} 

\REQUIRE{Criterion $\psi_{\#}(\bm{\theta},\mathbf{u})=\tilde{\psi}_{\#}(\bm{\theta},\mathbf{u})+\ln|\mathbf{K}_{\bm{\theta}}|$
to minimise for $(\mathbf{u},\bm{\theta})\in\mathbb{R}^{n}\times\mathbb{R}_{+}^{M}$.} 

\REPEAT

\STATE{Newton $\min_{\mathbf{u}}\psi_{\#}$ for fixed $\bm{\theta}$
(optional; few steps).}

\STATE{Refit upper bound: $\bm{\lambda}\leftarrow\nabla_{\bm{\theta}}\ln|\mathbf{K}_{\bm{\theta}}|=[\text{tr}(\mathbf{K}_{\bm{\theta}}^{-1}\mathbf{K}_{1}),..,\text{tr}(\mathbf{K}_{\bm{\theta}}^{-1}\mathbf{K}_{M})]^{\top}$.}

\STATE{Compute joint Newton search direction $\mathbf{d}$ for $\psi_{\bm{\lambda}}:=\tilde{\psi}_{\#}+\bm{\lambda}^{\top}\bm{\theta}$:
\\ \qquad{}\qquad{}$\nabla_{[\bm{\theta};\mathbf{u}]}^{2}\psi_{\bm{\lambda}}\mathbf{d}=-\nabla_{[\bm{\theta};\mathbf{u}]}\psi_{\bm{\lambda}}$.}

\STATE{Linesearch: Minimise $\psi_{\#}(\alpha)$ i.e. $\psi_{\#}(\bm{\theta},\mathbf{u})$
along $[\bm{\theta};\mathbf{u}]+\alpha\mathbf{d}$, $\alpha>0$.}

\UNTIL{Outer loop converged}

\end{algorithmic} \caption{\label{alg:double-loop} Double loop algorithm for joint MAP, MKL
and VB.}
\end{algorithm}

The Newton direction costs $O(n^{3}+M\, n^{2})$, with $n$ the number
of data points and $M$ the number of base kernels. All algorithms
discussed in this paper require $O(n^{3})$ time, apart from the requirement
to store the base matrices $\mathbf{K}_{m}$. The convergence proof
hinges on the fact that $\phi$ and $\phi_{\bm{\lambda}}$ are tangentially
equal \citep{nickisch09convexInf}. Equivalently, the algorithm can
be understood as Newton's method, yet dropping the part of the Hessian
corresponding to the $\ln|\mathbf{K}|$ term (note that $\nabla_{(\mathbf{u},\bm{\theta})}\phi_{\bm{\lambda}}=\nabla_{(\mathbf{u},\bm{\theta})}\phi$
for the Newton direction computation). Exact Newton for MKL.

In practice, we use $\mathbf{K}_{\bm{\theta}}=\sum_{m}\theta_{m}\mathbf{K}_{m}+\varepsilon\mathbf{I},\:\varepsilon=10^{-8}$
to avoid numerical problems when computing $\bm{\lambda}$ and $\ln|\mathbf{K}_{\bm{\theta}}|$.
We also have to enforce $\bm{\theta}\succeq\mathbf{0}$ in algorithm
\ref{alg:double-loop}, which is done by the barrier method \citep{Boyd:02}.
We minimise $t\phi+\mathbf{1}^{\top}(\ln\bm{\theta})$ instead of
$\phi$, increasing $t>0$ every few outer loop iterations.

A variant algorithm \ref{alg:double-loop} can be used to solve VB
in a different parametrisation ($\bm{\gamma}\succ\mathbf{0}$ replaces
$\mathbf{u}$), which has the same convexity structure as joint MAP.
Transforming equation \eqref{eq:vb-int} similarly to equation \eqref{eq:mlm-gauss},
we obtain 
\begin{equation}
\phi_{\text{VB}}(\bm{\theta})=\min_{\bm{\gamma}\succ\mathbf{0}}\ln|\mathbf{C}|-\ln|\bm{\Gamma}|+\bm{\beta}^{\top}\bm{\Gamma}\mathbf{C}^{-1}\bm{\Gamma}\bm{\beta}-\bm{\beta}^{\top}\bm{\Gamma}\bm{\beta}+h(\bm{\gamma})\label{eq:phi-comp}
\end{equation}
with $\mathbf{C}:=\mathbf{K}_{\bm{\theta}}+\bm{\Gamma}$, computed
using the Cholesky factorisation $\mathbf{C}=\mathbf{L}\mathbf{L}{}^{\top}$.
They cost $O(M\, n^{3})$ to compute, which is more expensive than
for joint MAP or MKL. Note that the cost $O(M\, n^{3})$ is not specific
to our particular relaxation or algorithm e.g. the Laplace MLM approximation
\citep{Williams:98b}, solved using gradients w.r.t. $\bm{\theta}$
only, comes with the same complexity.

\section{Conclusion\label{sec:conc}}

We presented a unifying probabilistic viewpoint to multiple kernel
learning that derives regularised risk approaches as special cases
of approximate Bayesian inference. We provided an efficient and provably
convergent optimisation algorithm suitable for regression, robust
regression and classification.

Our taxonomy of multiple kernel learning approaches connected many
previously only loosely related ideas and provided insights into the
common structure of the respective optimisation problems. Finally,
we proposed an algorithm to solve the latter efficiently.

\noindent \bibliography{mkl}

\end{document}